\documentclass[10pt, conference, compsocconf]{IEEEtran}
\usepackage{float}
\usepackage{booktabs}
\usepackage{caption}
\usepackage{multirow}
\usepackage{graphicx}
\usepackage{caption}
\usepackage{subcaption}
\usepackage{stfloats}
\usepackage{float}

\ifCLASSINFOpdf

\else
  
\fi

\begin{document}

\title{Exploring Multimodal Sentiment Analysis via CBAM Attention and Double-layer BiLSTM Architecture}


\author{
    \IEEEauthorblockN{
        Huiru Wang\IEEEauthorrefmark{2},
        Xiuhong Li\IEEEauthorrefmark{1},
        Zenyu Ren\IEEEauthorrefmark{3},
        Dan Yang\IEEEauthorrefmark{2},
        chunming Ma\IEEEauthorrefmark{2}
        \IEEEauthorblockA{
            \IEEEauthorrefmark{2}Xinjiang Key Laboratory of Signal Detection and Processing, College of Information Science and Engineering \\
            Xinjiang University, Urumqi, China\\
            Email: w176150@stu.xju.edu.cn; yangdan0223@stu.xju.edu.cn; mcm@stu.xju.edu.cn    
        } 
        \IEEEauthorblockA{
            \IEEEauthorrefmark{3}Xinjiang Multilingual Information Technology Laboratory, College of Information Science and Engineering \\
            Xinjiang University, Urumqi, China\\ 
            Email: renzeyu@stu.xju.edu.cn
        }
         \IEEEauthorblockA{
            \IEEEauthorrefmark{1}Xinjiang Key Laboratory of Signal Detection and Processing, College of Information Science and Engineering\\
            Xinjiang University, Urumqi, China\\ 
            Email:  xjulxh@xju.edu.cn
        }
    }
}


%


\maketitle

\begin{abstract}
Because multimodal data contains more modal information, multimodal sentiment analysis has become a recent research hotspot. However, redundant information is easily involved in feature fusion after feature extraction, which has a certain impact on the feature representation after fusion. Therefore, in this papaer, we propose a new multimodal sentiment analysis model. In our model, we use BERT + BiLSTM as new feature extractor to capture the long-distance dependencies in sentences and consider the position information of input sequences to obtain richer text features. To remove redundant information and make the network pay more attention to the correlation between image and text features, CNN and CBAM attention are added after splicing text features and picture features, to improve the feature representation ability. On the MVSA-single dataset and HFM dataset, compared with the baseline model, the ACC of our model is improved by 1.78\% and 1.91\%, and the F1 value is enhanced by 3.09\% and 2.0\%, respectively. The experimental results show that our model achieves a sound effect, similar to the advanced model.
\end{abstract}


%
\IEEEpeerreviewmaketitle

\section{Introduction}
With the advancement of information technology and the rapid development of the current short video social software platform, video software based on Douyin and Kuaishou has gradually become the primary platform for people to share their daily lives and express their opinions. Video is a type of multimodal data that typically includes sound, image, and other modalities. Video contains more information than text, and effective mining, fusion, and analysis of its various modalities can better understand video's views, which is of great value and significance to enterprise marketing and network public opinion analysis.

Social networks are primarily made up of multimodal data, the most important three of which are text, images, and voice. These three modalities collaborate and complement one another, presenting a good and complete multimodal data platform for users to express personal emotions.In the emotion analysis task, the human expression of emotions is also multimodal; for example, when only observing the text cannot accurately judge the speaker's situation, when the image with a smiley face and text combined with a high probability can be analyzed is a positive emotion. Therefore, it is necessary to use multimodal technology to analyze and mine multimodal emotions, whether in social media or in real life.

The use of multimodal data to classify target sentiment categories is referred to as multimodal sentiment analysis. The primary difference between multimodal sentiment analysis and single-modal sentiment analysis is whether the data input to the model is multiple modal information.  The primary goal of the multimodal sentiment analysis task is to combine information from multiple modalities in order to improve sentiment analysis accuracy. Compared to unimodal sentiment analysis, multimodal sentiment analysis presents a number of new challenges.

In general, information from different modalities can be complementary, meaning that information from more than one modality may provide more information than information from one modality. However, in most cases, modal fusion may contain redundant information of interference. It is not that the richer the fusion information, the higher the accuracy of sentiment analysis. Different modal information has different influences on the final emotional analysis results of the model. When multiple modes are similar emotional common characteristics, then the higher the possibility of accurately judging the polarity of the final unified emotion\cite{xu2017analyzing}. Complex judgment analysis will be performed if multiple modalities have different emotional characteristics. 
 
Recent sentiment analysis research has differed from previous studies in that it focuses on using deep learning-related techniques and multiple modalities to improve sentiment analysis task outcomes. After entering the multimodal data, extract the graphic and text features, obtain each modality's representation information, and then choose the appropriate multimodal data fusion algorithm to fuse various modal information to obtain a multimodal representation.

In this paper, our main contributions are as follows:

\begin{itemize}
    \item In order to achieve the fusion of different modal emotion features, we need to select the most applicable feature extraction methods for different data types, obtain as many emotion-related features as possible, extract and model the features of images and text separately, and use fusion methods to reduce the interference of irrelevant information.
    \item To better extract the sentiment features in the text, we use BERT+BiLSTM structure to enhance the feature extraction ability.
    \item In order to improve the feature expression ability after feature splicing, we added CBAM attention after CNN to pay more attention to modal related information.
\end{itemize}
This paper focuses on the current state of research in single-modal sentiment analysis, graphic sentiment analysis, and existing technology for integrating information from multiple modalities. To achieve the fusion of emotional features from different modalities, it is necessary to select the most applicable feature extraction method for different data types to obtain as many emotion-related features as possible, extract and model the features of images and texts respectively, and use the fusion method to reduce the interference of irrelevant information.

\section{RELEATE WORK}
The most common method in multimodal sentiment analysis is to analyze pictures and texts, that is, graphic sentiment analysis. Feature extraction technology and model training methods based on deep learning are often used in graphic sentiment analysis.

In the emotional analysis of images and texts, the first step is to choose the appropriate feature fusion method to fuse image features and text features. There are two standard feature fusion methods: feature-level fusion and decision-level fusion. Feature-level fusion is the process of combining image and text features and using them for model training and prediction. The goal of decision-level fusion is to combine the predictions of images and text to produce the final sentiment analysis results. In multimodal sentiment analysis, deep learning models such as convolution neural network and recurrent neural network are often used to realize joint modeling of images and texts.

Zhu et al. \cite{9736584} propose an innovative image-text interaction network (ITIN) that analyzes emotions expressed in social media posts by leveraging the interaction between images and text. They specifically intend to investigate the relationship between emotive picture areas and text in order to better understand the sentiment expressed in a given post. A convolutional neural network (CNN) is used to extract visual features from images, and a recurrent neural network (RNN) is used to process text content. After that, the two networks are combined in a joint architecture that learns to associate visual and textual features in order to predict the sentiment expressed in the post.Yang et al. \cite{9783103} propose a novel approach for sentiment analysis called TPMSA (Two-Phase Multi-Task Sentiment Analysis), in which they use a two-phase training technique and a novel multi-task learning strategy to improve classification performance. TPMSA's first phase entails pre-training a deep neural network with a large amount of unlabeled data. In the second phase, this pre-trained network is fine-tuned using a smaller labeled dataset. The goal of this two-phase training technique is to maximize the benefits of pre-training while also allowing the model to adapt to the specific task of sentiment analysis.Ye et al. \cite{ye2022sentiment} propose a sentiment-aware multimodal pre-training (SMP) sentiment analysis approach that focuses on extracting fine-grained sentiment information from sentiment-rich datasets. SMP aims to improve the performance of sentiment analysis models by combining textual and visual data. SMP entails pre-training a deep neural network on a large amount of unlabeled data containing both textual and visual information. Xiao et al. \cite{9747542} propose a novel approach for sentiment analysis called MAGCN (Multi-modal Attentive Graph Convolutional Networks) that uses the self-attention mechanism and densely connected graph convolutional networks to learn inter-modality dynamics. MAGCN entails creating a graph that connects the textual and visual characteristics of each sample in the dataset. The graph is densely connected, which means that each node is connected to every other node, and each edge is weighted based on the similarity of the connected nodes' features.Tang et al. \cite{9932611} propose the multimodal dynamic enhanced block to capture intra-modality sentiment context, and the bi-direction attention block to capture fine-grained multimodal sentiment context, using a novel bi-direction multimodal dynamic routing mechanism that allows us to dynamically update and investigate high-level and much finer-grained multimodal sentiment contexts. Zhao et al. \cite{zhao2019image} propose an image-text consistency measure for image-text postings that investigates the relationship between the picture and the text, which is then followed by a multimodal adaptive sentiment analysis approach.

Although multimodal sentiment analysis techniques are rapidly evolving, they still face many challenges and the accuracy of the models needs to be improved. How to effectively utilize more inter-modal correlation information and remove redundant information will be the focus of this paper.
\section{MATERIALS AND METHODS}
Based on the previous work \cite{wang2023multimodal} (Multilayer fusion convolutional neural networks (MLFC) and Supervised Comparative Learning), we propose a new multimodal sentiment analysis model. This model is effective in solving the problem of interference information redundancy in the process of modal fusion. In addition, although CNN has extracted local features and some redundant information is removed in feature fusion, we still hope to obtain more feature information and suppress the influence of irrelevant information on the model.

Although BERT pre-training model has achieved remarkable results in text feature extraction tasks, there is still room for improvement. To obtain rich feature information in the text, we add a double-layer BiLSTM structure to the pre-training model BERT, which can capture the context information in the text more comprehensively and improve the expressive ability of the text. Moreover, this improvement can also promote the deep fusion of modal information and enhance the strong correlation between text features and image features. 

The BERT+double-layer BiLSTM method can improve text feature extraction and model accuracy and effect in multimodal sentiment analysis tasks. Moreover, it can make up for the missing position information in BERT, capture the long-distance dependency relationship in sentences, and learn the feature representation of different levels of input sequences, which is helpful for the model to better understand the semantics of sentences.

\begin{figure*}
\includegraphics[width=\textwidth]{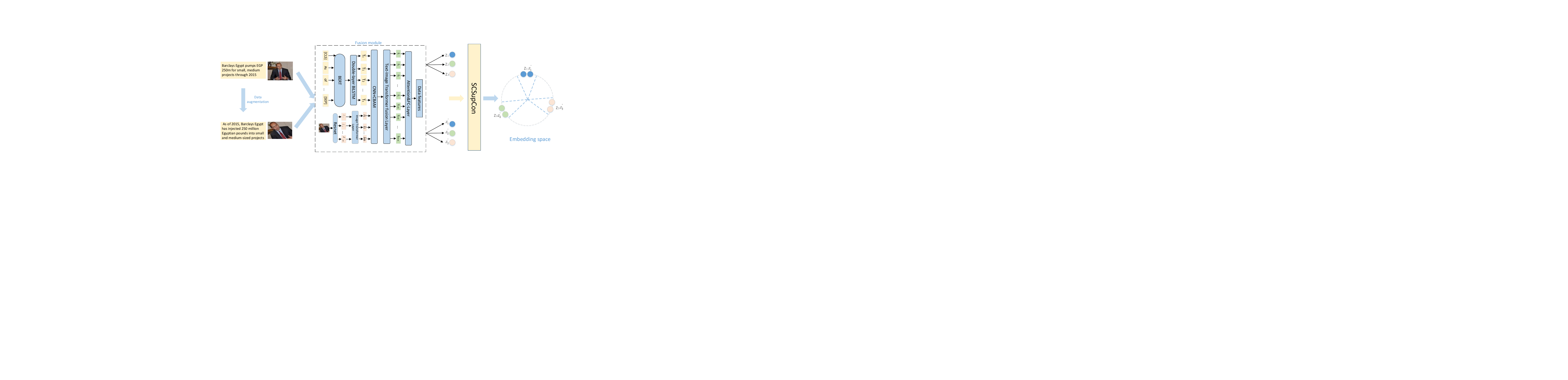}
\caption{Illustration of our Model overall framework diagram.To judge sentiment polarity, the~proposed architecture employs supervised contrastive learning and a CNN-connected Transformer fusion. The proposed architecture adopts supervised comparative learning and transformer fusion of CNN and CBAM connections. Invariance occurs for graphic sample features when embeddings from the same category, such as $z_{1}$ and $z'_{1}$. Different categories of graphic features, such as $z_{1}$ and $z_{2}$, on~the other hand, are far apart.}
\label{fig1}
\end{figure*}
\unskip
Adding Convolutional Block Attention Module (CBAM) attention\cite{woo2018cbam} after CNN can effectively capture the correlation between spatial and channel dimensions, learn features with different granularity, and improve the performance of the model without increasing too much computational cost. To sum up, adding the CBAM attention module can effectively improve the performance and performance of the CNN model. Figure~\ref{fig1} shows the overall structure of our proposed model method, which is mainly divided into four modules.
\subsection{Feature Extractor}
For sentiment analysis task, obtaining sufficient unimodal information is an important element. Feature extraction is an important way to obtain unimodal information. Resnet-50 and vanilla Transformer\cite{vaswani2017attention} are chosen as image feature extractors. And BERT pre-trained models are chosen as text feature extractors. 

The model inputs a set of text image data samples and at the same time generates enhanced samples through data augmentation, and both sets of data pass through the fusion network structure.

To obtain the text hidden representation, we feed the text encoding output by the BERT pre-trained model into double-layer BiLSTM. Forward and backward LSTM are used to calculate the output of the input word embedding vector. The first layer's output is used as the input for parameter training in the second layer BiLSTM, to get the final text feature output.
For convenience, we use the symbol $t$ to represent the text input.

We let the BERT pre-trained model and double-layer BiLSTM as our text feature extractor:
\begin{equation}
     T=Extractor(t)
\end{equation}

Where, $T=\{t_{c},t_{1},t_{2},...,t_{s}\}$ refer to the word embedding vector of the extractor output. $T\in R^{n_{t}*d_{t}}$

The original image $m$ is fed into ResNet-50 and the output of the last layer of the ResNet-50 convolutional layer is used as a hidden representation of the image features $M_c$:
\begin{equation}
    M_c=ResNet(m)
\end{equation}

We convert this hidden representation dimension to the same dimension as the text $T$:
\begin{equation}
    M_1=flatten(M_{c}W_{M}+b_{M})
\end{equation}

We can get the sequence feature representation of the image 
$M_{1}=\left\{m'_{1},m'_{2},\cdots,m'_{n_i}\right\}$, $M_{1}\in{R}^{n_i\times d_t}$,and input this sequence feature representation into the Transformer to get the global feature $M$ of the image.
\subsection{Fusion Framework}

The image hidden encoding and text feature representation obtained from feature extraction are stitched and input to the fusion layer. Multimodal data is fused by CNN and Transformer to remove interference and redundant information of modal fusion, but in this fusion mode, the graphic sequence processed by CNN lacks modal correlation. Therefore, we add CBAM attention after the CNN to better focus on the sentiment-related information and cooperate with the transformer to achieve global feature fusion. The following is the formula:
\begin{equation}
     \{f_{1},f_{2},...,f_{{n_{t}+n_{i}}\}=Transformer\{CBAM\{CNN[concat(T,M)]\}}
\end{equation}
\begin{equation}
     F'=\{f_{1},f_{2},...,f_{n_{t}+n_{i}}\}
\end{equation}

We obtain the multimodal fusion graphic sequence representation $F'$ via the fusion layer, but the generated graphic sequence cannot be classified directly, so we must use the fully connected layer and a simple attention layer to obtain the multimodal representation $R'$. The output contains results that can be used to perform classification tasks.
\begin{equation}
     R'=Attention(F')
\end{equation}
\subsection{Loss of Sentiment Classification}

Supervised comparative learning enables the model to better learn the common features in the data by comparing positive and negative samples for learning, thus improving its generalization performance. The model can be assisted to learn the common features of the same characteristics using contrast learning, but there are false ``positive examples'' and too few positive examples. We use supervised contrast learning to obtain more positive examples and solve the problem of false ``positive examples''. The more positive and positive samples there are, the more the model learns about the characteristics of affective tendencies and improves the accuracy of the model analysis. A loss function, which defines the penalty for the model's failure to predict, is first defined in contrastive learning. This function is then used by the machine learning program to adjust the model parameters in order to minimize the loss value, enabling accurate supervised learning. The expression for the supervised contrast learning loss function as follow:
\begin{equation}
     L_{SupCon}=\sum_{i=1}^{n}\frac{-1}{|P(i)}\sum_{p\in P(i)}log\frac{exp(z_{i}\cdot z_{p}/\tau)}{\sum_{a\in A(i)}exp(z_{i}\cdot z_{a}/\tau)}
\end{equation}

The supervised comparative learning loss function is used as our classification loss to solve the problem of too few positive samples and false negative samples when the sample label is known. It obtain more positive samples for supervised comparative learning tasks, so that the model can learn more common features of emotions, and improve the accuracy of model sentiment analysis. Use a cross-entropy loss function to calculate the difference between the model's predicted and true outcomes.
\begin{equation}
     L_{SC}=Cross-Entropy(GELU(RW_{sc}+b_{sc}))
\end{equation}
\begin{equation}
    L_{SCSupCon}=\lambda_{sc}L_{SC} + \lambda_{Supcon}L_{SupCon}
\end{equation}
\section{EXPERIMENGTS}
To achieve the purpose of multimodal sentiment analysis, the two different modal data of the two datasets are fused after feature extraction, and then supervised comparative learning is used on the fused modal features, so that the closer the similar features in the feature space, the better, and the different types of features are far away from each other.

We built the proposed model using the Pytorch framework, with ACC and F1 as the primary evaluation metrics. Our research was conducted on a high-performance computing (HPC) node with a taitanrtx GPU. The parameters of the pre-trained BERT model and the Resnet-50 model are fixed, while the parameters of word embedding and attribute embedding are constantly updated during the training process.
\subsection{Evaluation Indicators}
In the multimodal sentiment analysis task, it is important to accurately evaluate the performance of the model, so it is necessary to choose the appropriate evaluation indicators to measure the performance of the model. This article uses the two most commonly used evaluation metrics: accuracy rate (ACC) and F1 score. We use accuracy and f1 value as the measurement criteria for model evaluation indicators.

The F1 score is a popular evaluation metric. It combines accuracy and recall to assess a classifier's performance.

To more accurately assess the model's performance in a multi-class classification problem, the accuracy and F1 score for each category can be calculated\cite{yang2021multimodal}. Sentiment categories can be treated as categorical labels in multimodal sentiment analysis tasks, and the accuracy and F1 score for each sentiment category are calculated. 
\subsection{Data preprocessing}

The hyperparameter setting is back translation for text, RandAugment for images, the emotional polarity of the MVSA-Single dataset is 3, the emotional polarity of the HFM dataset is 2, the contrastive learning method is supervised contrastive learning, epoch is set to 50, and the optimizer is adamw.

The original two MVSA datasets are processed in the same manner as described in the \cite{xu2017multisentinet}, and the MVSA datasets are randomly divided into training, validation, and test sets with an 8:1:1 segmentation ratio. The same data preprocessing methods as in the literature\cite{cai2019multi} were used for HFM. Along with randomly dividing the dataset, manually check the development and test sets to ensure label accuracy. The most common method is to employ the NLTK toolkit\cite{bird2006nltk}.

\subsection{Comparative Experiments}
We compare and analyze our model with other models in the multimodal sentiment analysis task. The models selected for comparison are as follows:

SentiBank + SentiStrength$:$ SentiBank can be used to extract multiple adjective-noun pairs from images, and SentiStrength can be used to determine the text's emotional polarity\cite{borth2013large}.

CBOW+DA+LR$:$ Unsupervised learning is used to acquire visual information from a large-scale corpus, which is then combined with a neural network-based language model for multimodal sentiment analysis\cite{baecchi2016multimodal}.

CNN-Multi$:$ Text and image features are extracted using two CNN models for layered fusion, and these features are used to analyze sentiment\cite{cai2015convolutional}.

DNN-LR$:$ Text and image features are extracted using two CNN models, and these features are used to analyze sentiment using logistic regression\cite{yu2016visual}.

MN-Hop$:$ The attention mechanism is used to interactively model the relationship between text and visual memory. MN-Hop2+ img2text is a variant of MN-Hop that proposes a co-memory network to iterate the interaction between text and images to learn to influence each other\cite{xu2018co}.

CFF-ATT$:$ The denoising autoencoder is used to remove the effect of noise in the text, the attention-based variational autoencoder is used to extract image features, and the internal text and image features are learned symmetrically with each other\cite{zhang2020sentiment}.

MVAN-M$:$ This is the use of memory networks that are constantly updated to obtain deep semantic information about text and images\cite{yang2020image}.

MFB$:$To jointly learn images and problem attention, the multimodal decomposition bilinear pool method combines multimodal features and co-attention mechanisms\cite{yu2017multi}.

FENET-Glove and FENET-BERT$:$ To effectively extract the relationship and influence between text and image, a fine-grained attention mechanism is used. Furthermore, a new method based on a gating convolution mechanism is being investigated for extracting and utilizing visual and textual information for sentiment prediction\cite{jiang2020fusion}.

CMCN$:$The use of cross-modal complementary networks and hierarchical synthesis in multimodal sentiment analysis tasks. The model is organized into three major modules, resulting in a multi-level fusion system that allows for the full integration of various modal features while reducing the risk of integrating irrelevant modal features\cite{peng2021cross}.

CLMLF$:$ Contrastive learning and multi-layer fusion methods for multimodal sentiment detection. Firstly, the text and image are encoded to obtain a hidden representation, and then the multi-layer fusion module is used to align and fuse the token-level features of the text and image, and two contrastive learning tasks, label-based contrastive learning and data-based contrastive learning task, are also designed\cite{li2022clmlf}.

Schifanella et al. concatenate different feature vectors of different modalities into multimodal feature representations for the HFM dataset, batch we set to 24. Concat(2) connects text and image features, whereas Concat(3) adds another image attribute feature\cite{schifanella2016detecting}.

We conducted relevant experiments on the small sample dataset MVSA-Single and the larger dataset HFM to demonstrate the effectiveness of the model. The results show that our model is comparable to the SOTA model in terms of effectiveness. The following table \ref{TableD} and table \ref{TableE} show the detailed results of the performance metrics.

\begin{table}[!ht]
    \centering
    \caption{\textbf{MVSA dataset comparison experiment}}
    \label{TableD}
    \begin{tabular}{ccc}
    
    \toprule
        Model & ACC & F1 \\ 
        \midrule
        SentiBank + SentiStrength & 52.05\% & 50.08\% \\ 
        CBOW+DA+LR & 63.86\% & 63.52\% \\ 
        CNN-Multi & 61.20\% & 58.37\% \\ 
        DNN-LR & 61.42\% & 61.03\% \\ 
        MN-Hop2+ img2text & 68.07\% & 65.19\% \\ 
        CFF-ATT & 71.44\% & 71.06\% \\
        MFB & 71.62\% & 72.98\% \\
        MVAN-M & 72.98\% & 72.98\% \\
        FENET-Glove & 72.54\% & 72.32\% \\
        FENET-BERT & 74.21\% & 74.06\% \\
        CMCN & 73.61\% & 75.03\% \\
        CLMLF & 75.33\% & 73.46\% \\
        Ours & 77.11\% & 76.55\% \\
    \bottomrule
    \end{tabular}
\end{table}
Our evaluation metrics ACC and F1 improved by 1.78\% and 3.09\%, respectively, when compared to the CLMLF model, as shown in the table\ref{TableD}. Overall, our model outperforms other models on the MVSA-Single public dataset, demonstrating the study's efficacy in multimodality and various emotional polarities. We compare various models to demonstrate the applicability of the models in this chapter, solve the feature fusion problem between modalities to some extent, and learn more emotional features.
\begin{table}[!ht]
    \centering
    \caption{\textbf{HFM dataset comparison experiment}}
    \label{TableE}
    \begin{tabular}{ccc}
    
    \toprule
        Model & ACC & F1 \\ 
        \midrule
        Concat(2) & 81.03\% & 77.99\% \\ 
        Concat(3) & 81.74\% & 78.74\% \\ 
        MMSD & 83.44\% & 80.81\% \\ 
        D\&R Net & 85.02\% & 80.60\% \\ 
        CLMLF & 85.43\% & 84.87\% \\ 
        Ours & 87.34\% & 86.87\% \\
    \bottomrule
    \end{tabular}
\end{table}

As shown in the table\ref{TableE}, we compared the HFM dataset to five related multimodal models, and the experimental results show that the model in this chapter improves the evaluation index ACC by 1.91\% and the F1 value by 2.0\% when compared to the CLMLF model. In general, because the hardware environment is low for HFM datasets with large sample data, it is impossible to set a higher batch, but our model relies more on larger batch conditions due to supervised comparative learning, and the overall improvement effect is better than that of MVSA-Single with a small sample dataset.

\subsection{Ablation Experiments}
We conducted ablation experiments on MVSA-single and HFM datasets to demonstrate the necessity of CBAM attention and double-layer BiLSTM in our model.

\begin{table*}[htb]
    \centering
    \caption{\textbf{MVSA-Single and HFM dataset ablation experiments}}
    \label{TableC}
    \tabcolsep=0.016\linewidth
    \begin{tabular}{ccccccc}
    \toprule
        \multirow{2.5}{*}{Method} & \multicolumn{2}{c}{MVSA-Single}  & \multicolumn{2}{c}{HFM} \\
        \cmidrule{2-5}
        & ACC & F1 & ACC & F1 \\
         \midrule
    BERT & 71.11\% & 69.70\% & 83.89\% & 83.26\%  \\
    ResNet-50 & 64.67\% & 61.55\% & 72.77\% & 71.38\% \\ 
    BERT+ ResNet-50+MLF+DBCL+LBCL & 75.33\% & 73.46\% & 85.43\% & 84.87\%  \\ 
    BERT+ ResNet-50+MLF+SCSupConLoss & 75.33\% & 75.75\% & 86.27\% & 85.70\% \\ 
    BERT+ ResNet-50+MLFC+DBCL+LBCL & 76.00\% & 75.11\% & 86.35\% & 85.99\% \\ 
    BERT+ ResNet-50+MLFC+ SCSupConLoss & 76.44\% & 75.61\% & 86.64\% & 86.22\% \\ 
    BERT+ ResNet-50+MLFC+CBAM+ SCSupConLoss & 76.89\% & 76.06\% & 86.85\% & 86.38\% \\ 
    BERT+Double-layer BiLSTM+ResNet-50+MLFC+CBAM+ SCSupConLoss & 77.11\% & 76.55\% & 87.34\% & 86.87\% \\
\bottomrule
\end{tabular}
\end{table*}

The table\ref{TableC} shows that when the contrastive learning method and MLF module are used, the model's performance decreases, and adding Sentiment Classification Supervised Contrastive Loss (SCSupConLoss) to the ablation experiment results shows that it improves the contrast learning effect. The final experimental results show that the addition of two modules improves the model's utility, and the addition of SCSupConLoss combined with Multi-Layer Fusion Convolution Neural Network (MLFC) aids in the extraction of local features and improves the model's overall performance. All experimental indicators were improved, demonstrating the model's effectiveness.

As shown in the table\ref{TableC}, CBAM can effectively direct the fusion process to focus on important information while suppressing irrelevant data, which has been improved in ACC and F1. On this basis, we augment the text feature extraction with a Bidirectional Long Short Term Memory (BiLSTM) to obtain more text feature information and mine deeper semantic information. It is possible to demonstrate that our model produced positive results.

Based on previous related work, we added the attention mechanism in the multimodal fusion part, which improved the ACC and F1 evaluation by about 0.5\% each, and we added the two-layer BiLSTM structure of text feature extraction, which improved by about 0.8\%. Based on previous related work, we add an attention mechanism and text feature extraction bilayer BiLSTM structure to the multimodal fusion part for HFM datasets, which improves by about 0.7\%.

\section{Conclusion}
This paper proposes a new multimodal sentiment analysis model that uses BERT and BiLSTM as feature extractors to mine rich emotional features in text. In order to reduce the interference information and enhance the network's attention to the correlation information between modes, CNN and CBAM attention mechanisms are added after the stitching of text features and image features to improve the feature representation ability. Experimental results show that our model achieves similar performance to advanced models.

The sentiment analysis method based on supervised contrastive learning and multimodal fusion will continue to have a broad research space and development prospects in the future. Future research could look into how adversarial learning and knowledge graphs can be used to improve the performance and robustness of sentiment analysis. With the advancement of globalization, an increasing number of people require cross-language sentiment analysis. Future research can look into how to improve the accuracy and robustness of cross-lingual sentiment analysis by combining multilingual sentiment analysis with methods based on supervised contrastive learning and multimodal fusion. Finally, the sentiment analysis method based on supervised contrastive learning and multimodal fusion has a broad application potential, and future research can be expanded and improved in many ways.

\section*{Acknowledgment}

This research was supported by the open project of key laboratory, Xinjiang Uygur Autonomous Region (No. 2022D04079) and the National Science Foundation of China (No. 11504313, No. U1911401, No. 61433012)

\bibliographystyle{IEEEtran}

\bibliography{RA}

\end{document}